\documentclass{llncs}
\usepackage{amsmath,graphicx}
\usepackage{amsmath,amssymb} 
\usepackage{epsfig}
\usepackage{multirow}
\usepackage{llncsdoc}
\usepackage{etoolbox}

\begin{document}

\mainmatter

\title{Retinal Vasculature Segmentation Using Local Saliency Maps and Generative Adversarial Networks For Image Super Resolution}

\titlerunning{Super Resolution using GANs and Saliency Maps}

\author{Dwarikanath Mahapatra, Behzad Bozorgtabar, Sajini Hewavitharanage, Rahil Garnavi
}


\authorrunning{Mahapatra et al.}
\institute{IBM Research Australia \\
\email{[dwarim,sydb,sajinihe,rahilgar]@au1.ibm.com.}}

\maketitle

\begin{abstract}
We propose an image super resolution(ISR) method using generative adversarial networks (GANs) that takes a low resolution input fundus image and generates a high resolution super resolved (SR) image upto scaling factor of $16$. This facilitates more accurate automated image analysis, especially for small or blurred landmarks and pathologies. Local saliency maps, which define each pixel's importance, are used to define a novel saliency loss in the GAN cost function. Experimental results show the resulting SR images have perceptual quality very close to the original images and perform better than competing methods that do not weigh pixels according to their importance. When used for retinal vasculature segmentation, our SR images result in accuracy levels close to those obtained when using the original images.
\end{abstract}


\section{Introduction}
\label{sec:intro}

Normal retinal fundus images have high resolution to detect and segment prominent landmarks and pathologies, but not sufficient for small and indistinct pathologies (microaneurysms, haemorrhages) and smaller vessel branches. Image super resolution (ISR) produces highly accurate super resolved (SR) images from single field of view(FOV) images that are comparable with the original HR images.  This is particularly relevant for tele-ophthalmology requiring transmission of acquired LR images. SR images improves detection of neovascularization, segmentation of small vessel branches and small pathologies not visible in the original LR images. Ophthalmologist may also use it to closely analyze suspicious regions with minute abnormalities.


Medical ISR methods using examples \cite{Jog8} and self similarity \cite{Jog7} were reliant on external data which put them at a disadvantage. 
Subsequently, parametric generative models learned the mapping between the original and LR version 
to upscale MR brain \cite{Jog9} and cardiac \cite{Oktay3} images.
These approaches are computationally demanding as candidate patches are searched in the training dataset to find the most suitable HR sample.
Other methods using random  forest regressors \cite{TannoMICCAI16}  convolutional neural networks (CNNs) \cite{SRCNN} produce high quality images. Recent work using Fourier burst accumulation \cite{Jog}, generative adversarial networks (GANs) \cite{Srgan} and CNNs \cite{Oktay} also highlight the importance of ISR for medical image analysis problems.

GANs \cite{Srgan} are state-of-the-art for ISR primarily due to the ResNet based  generator architecture, but are less effective with retinal images for scaling factors greater than $4$ due to cost functions that do not explicitly include local structure information. To overcome this limitation we propose a image SR method based on GANs that has the following novelties in its cost function: 1) using local saliency maps computed from curvature maps (that highlight local structures) and 2) entropy filtering (to highlight compact regions). Together they outperform \cite{Srgan} for $4-16$x magnification and the resulting SR images preserve information content and perceptual information of the LR image.
Our method's effectiveness is demonstrated in segmenting the retinal vasculature of SR images.


\section{Saliency Map Calculation}
\label{met:saliency}

Existing saliency methods highlight a globally salient region while `local' saliency maps are essential to compute individual pixel importance values.
Inspired by Perazzi et al \cite{Perazzi_CVPR} we combining abstraction, element distribution and uniqueness to generate a local saliency map for retinal images.

\textbf{Abstraction:}
Using superpixels for abstraction (as in \cite{Perazzi_CVPR}) provides high level global information. Instead curvature maps ($I_{curv}$) capture local structural information based on vessel curvature and other small elements.
%
 %
 \begin{equation}
 I_{Curv}=\frac{f_{xx}f_y^{2} + f_{yy}f_x^{2} - 2f_{xy}f_xf_y}{\left(f_x^{2}+f_y^{2} \right) ^{3/2} },
 \end{equation}
 where $f_x,f_y$ are image gradients and $f_{xx},f_{yy}$ indicate second derivatives.

\textbf{Element Distribution:}
Element distribution captures compactness and continuity of retinal structures. 
Pixel ($s$) entropy is given by  $I_{Ent}(s)=-\sum_{i\in N_s} p_i \log p_i$. 
 $I_{Ent}$ is the entropy image, $N_s=7\times7$ is $s$'s neighborhood and determines compactness, $p_i$ is the probability of intensity $i$ in $N_s$ calculated using a $8$ bin histogram (experimentally determined). $I_{Ent}$ assigns low values for regions with compact objects. Hence its values are normalized to $[0,1]$ and transformed as $1-I_{Ent}$ to highlight compact regions. $I_{Ent}$ is smoothed using a Gaussian low pass filter of size $3$ and standard deviation $0.5$ to remove any isolated noisy regions or pixel clusters.

\begin{figure}[t]
\begin{tabular}{ccccc}
\includegraphics[height=1.99cm, width=2.3cm]{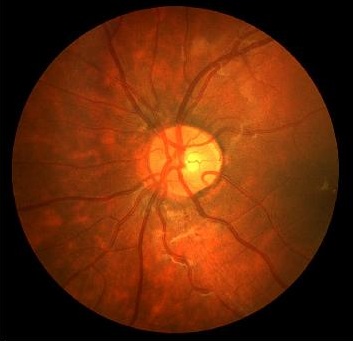}  & 
\includegraphics[height=1.99cm, width=2.3cm]{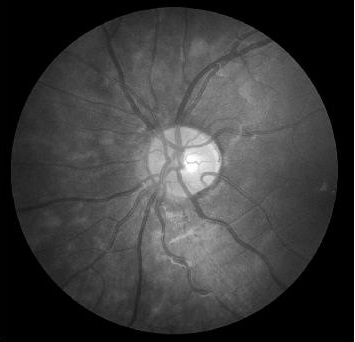}  & 
\includegraphics[height=1.99cm, width=2.3cm]{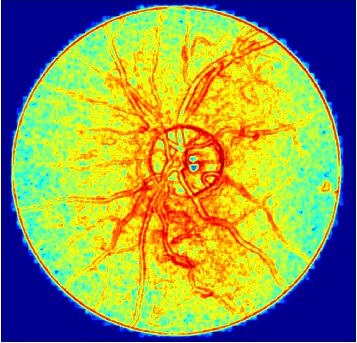}  & 
\includegraphics[height=1.99cm, width=2.3cm]{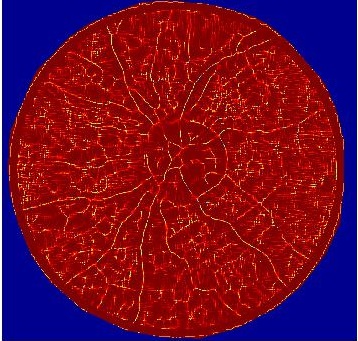}  & 
\includegraphics[height=1.99cm, width=2.3cm]{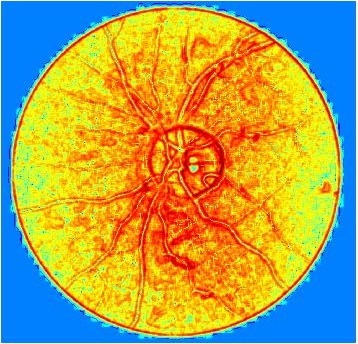}  \\ 
(a) & (b) & (c) & (d) & (e) \\
\end{tabular}
\caption{(a) Original fundus image (b) grayscale image; (c) $1-I_{Ent}$; (d) $I_{Curv}$; (e) local saliency map $I_{Sal}$ at scale $7\times7$. Warmer colours indicate higher values}
\label{fig:salmaps3}
\end{figure}

\textbf{Uniqueness:}
The sum of weighted difference of pixel feature maps is,
\begin{equation}
D_F(s)=\sum_i {\exp\left(-\left\|s-s_i\right\|\right)\left|F(s)-F(s_i)\right| }, 
\label{eqn:diffmap}
\end{equation}
where $D_F$ indicates the difference or uniqueness map for feature map $F$ ($I_{Curv}$ or  $1-I_{Ent}$); $\left\|s-s_i\right\|$ is the Euclidean distance between $s$ and its $i$th neighbor $s_i$. %
 \emph{Squared difference}  of feature values in \cite{Perazzi_CVPR} introduces blur which is undesirable for image super resolution. Instead we use the \emph{absolute difference} and also normalize the difference map to $[0,1]$. 
The final saliency map is, %
\begin{equation}
I_{Sal}=w_1\times D_{Curv} ~ + ~ (1-w_1)\times D_{1-Ent}.
\label{eqn:Salmap}
\end{equation}
where $w_1$ balances the relative contribution of each feature.   
$w_1=0.4$ was experimentally set by varying it between $[0,1]$ in steps of $0.01$ and comparing the quality of the resulting SR images on a subset of $50$ images. 
The resulting saliency map in Figure~\ref{fig:salmaps3} (e) clearly highlights the local retinal structures and thus justifies its use in the GAN cost function.  

\section{Generative Adversarial Networks}

\begin{figure}[t]
\begin{tabular}{ccccc}
\includegraphics[height=2.5cm, width=5.4cm]{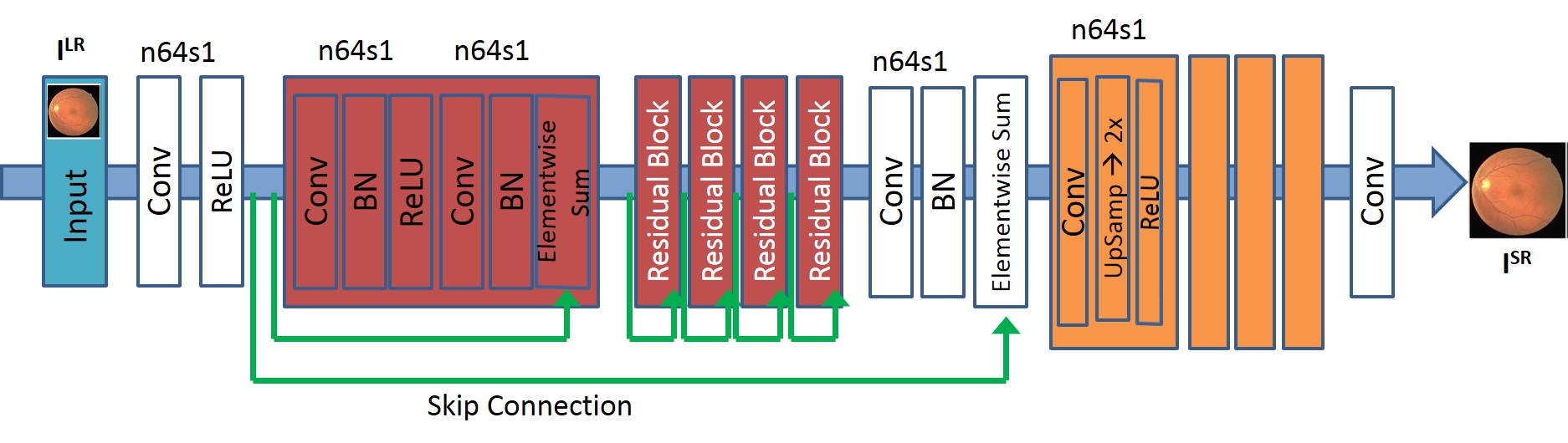}  &
& & &
\includegraphics[height=2.5cm, width=5.99cm]{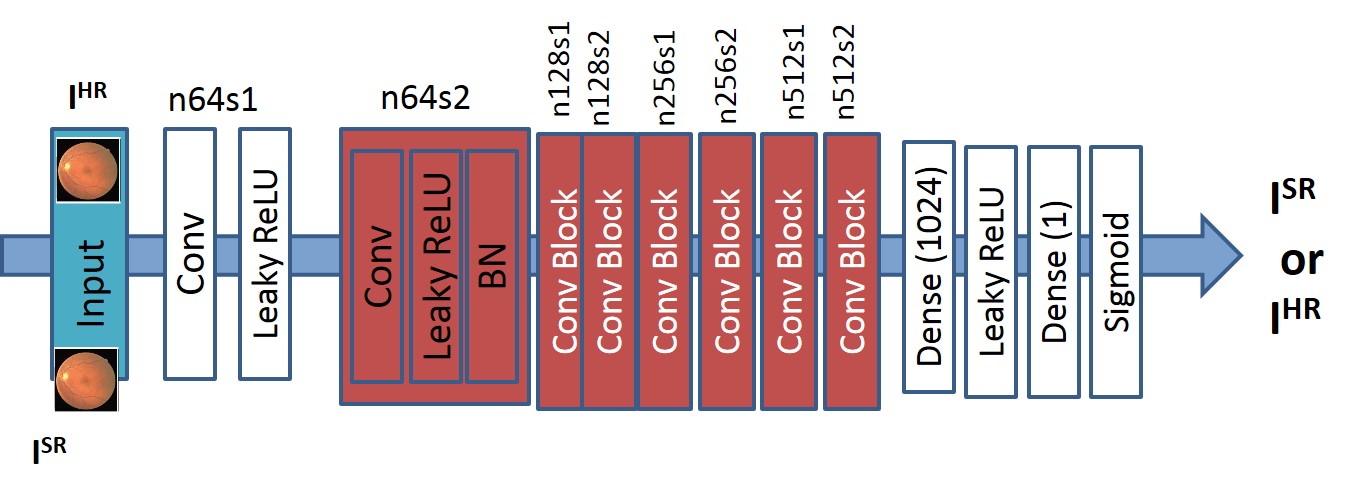}  \\
(a) & & & & (b)\\
\end{tabular}
\caption{(a) Generator Network; (b) Discriminator network. $n64s1$ denotes $64$ feature maps (n) and stride (s) $1$ for each convolutional layer. }
\label{fig:Gan}
\end{figure}

ISR estimates a high-resolution, super resolved image $I^{SR}$ from a low-resolution input image $I^{LR}$. For training, $I^{LR}$ is the low-resolution version of the high resolution counterpart $I^{HR}$, obtained by applying a Gaussian filter to $I_{HR}$ followed by downsampling with factor $\textbf{r}$. 
The generator network is a feed-forward CNN ($G_{\theta_G}$) whose parameters $\theta_G = {W_{1:L}; b_{1:L}}$ are obtained by, 
\begin{equation}
\widehat{\theta}=\arg \min_{\theta_G} \frac{1}{N} \sum_{n=1}^{N} l^{SR}\left(G_{\theta_G}(I_n^{LR}),I_n^{HR}\right),
\label{eq:theta1}
\end{equation}
where $l^{SR}$  is the loss function and $I_n^{HR}$, $I_n^{LR}$ are HR and LR images.
The adversarial min-max problem is defined by,
\begin{equation} 
\min_{\theta_G} \max_{\theta_D}  \mathop{\mathbb{E}}_{I^{HR}~p_{train}(I^{HR})}[\log D_{\theta_D}(I^{HR})]  + 
 \mathop{\mathbb{E}}_{I^{LR}~p_{G}(I^{LR})}[\log (1-D_{\theta_D}(G_{\theta_G}(I^{LR}))] 
\label{eq:cf1}
\end{equation}
This trains  a generative model $G$
 with the goal of fooling a
differentiable discriminator $D$ that is trained to distinguish SR images from real images. 
$G$ creates solutions that are very similar to real images and thus difficult to classify by $D$.
This encourages perceptually superior solutions than obtained by minimizing pixel-wise mean square error (MSE).
$G$ employs residual blocks (Figure~\ref{fig:Gan} (a)). Each block has two convolutional layers with $3\times3$ filters and $64$ feature maps,  followed by batch normalization and ReLU activation.

$D$  solves the maximization problem in Eqn.~\ref{eq:cf1}.
It has eight 
convolutional layers with the 
kernels increasing by a factor of $2$ from $64$ to $512$ (Figure~\ref{fig:Gan} (b)). Leaky ReLU is used and strided convolutions reduce the image dimension  when the number of
features is doubled. The resulting $512$ feature maps are
followed by two dense layers and a final sigmoid activation to obtain a probability map.

\subsection{Loss Function}

 $l^{SR}$ is a combination of content loss ($l_{Cont}^{SR}$) and adversarial or generative loss ($l_{Gen}^{SR}$), balanced by a factor $\alpha=0.01$, and is given by  : 
\begin{equation}
l^{SR} = l_{Cont}^{SR} + \alpha l_{Gen}^{SR}
\label{eq:cf2}
\end{equation}

\textbf{Content Loss:}
MSE results in smooth SR images that are perceptually unsatisfying and lack high frequency content. %
Perceptually important details in SR images is preserved by our saliency weighted MSE loss ($l_{w-MSE}$),
\begin{equation}
l_{w-MSE}=\frac{1}{WH} \sum _{x=1}^{W} \sum _{y=1}^{H} \times (w_I^{HR}I^{HR}_{x,y} -w_I^{SR}G_{\theta_G} (I^{LR})_{x,y} )^2,
\label{eq:wlMSE}
\end{equation}
where $w_I^{HR},w_I^{SR}$ are saliency values of HR ($I^{HR}$) and SR ($G_{\theta_G} (I^{LR})$) images.
%
A CNN loss \cite{Srgan} is calculated as the $L2$ distance between SR image and ground-truth HR image using all $512$ feature maps of Relu $4-1$ layer of a pre-trained $VGG-16$ \cite{VGG}. 
\begin{equation}
l^{SR}_{CNN} =\frac{1}{W_{i,j}H_{i,j}} \sum _{x=1}^{W_{i,j}} \sum _{y=1}^{H_{i,j}} (\phi_{i,j}(I^{HR})_{x,y} -\phi_{i,j}(G_{\theta_G} (I^{LR}))_{x,y} )^2
\label{eq:lCNN}
\end{equation}
$\phi_{i,j}$ the feature map obtained by the $j-$th convolution  (after activation) before the $i-$th max pooling layer and
$W_{i,j}$ and $H_{i,j}$ are the dimensions of $\phi$.

\textbf{Local Saliency Loss:} Our novel saliency loss measures the difference in saliency maps of $G_{\theta_G} (I^{LR}))$ and $I^{HR}$ by comparing their local landmarks. This enables inclusion of important structural information in the cost function.
\begin{equation}
l^{SR}_{Sal} =\frac{1}{W_{i,j}H_{i,j}} \sum _{x=1}^{W_{i,j}} \sum _{y=1}^{H_{i,j}} ((I^{HR}_{Sal})_{x,y} -(G_{\theta_G} (I^{LR})_{Sal})_{x,y} )^2
\label{eq:lSal}
\end{equation}
$I^{HR}_{Sal}$ and $G_{\theta_G} (I^{LR})_{Sal}$ denote the saliency maps of $I^{HR}$ and $G_{\theta_G}(I^{LR})$. %

\textbf{Adversarial Loss:}
The generative loss $l^{SR}_{Gen}$ \cite{Srgan} over all training samples is 
\begin{equation}
l^{SR}_{Gen}=\sum_{n=1}^N -\log D_{\theta_D}(G_{\theta_G}(I^{LR}))
\end{equation}
$D_{\theta_D}(G_{\theta_G}(I^{LR}))$  is probability that  $G_{\theta_G}(I^{LR})$ is a natural HR image. This network favours solutions in the manifold of retinal images. Convergence is facilitated by minimizing $-\log D_{\theta_D}(G_{\theta_G}(I^{LR}))$ instead of $-\log [1-D_{\theta_D}(G_{\theta_G}(I^{LR}))]$.

\textbf{Training:}
The model in Fig.~\ref{fig:Gan} (a) generates images upscaled by $2\times$. For higher scale factors we feed the output of this network to another identical network to get $4\times$ images. Consecutive networks can be combined to get $8\times$, $16\times$, etc image resolution. In each subsequent upsampling step the corresponding networks are trained on images of size $4\times$, $8\times$, etc.


\section{Experiments And Results}
\label{sec:expt}

\textbf{Dataset:}
We apply our algorithm on $5000$ retinal fundus images from multiple sources with different image dimensions \cite{data}, and augmented $100$ times by rotation and translation. The dark borders were removed and the images resized to $1024\times1024$ pixels. Our method was implemented with Python and TensorFlow (for GANs). For GAN optimization we use Adam with $\beta_1=0.93$ and batch normalization. The ResNet was trained with a learning rate of $0.001$ and $10^{5}$ update iterations. MSE based ResNet was used to initialize $G$. The final GAN was trained with $10^{5}$ update iterations at learning rate $10^{-3}$. 
The average training time using the augmented version from $4000$ images was $14$ hrs for scaling factor(r) $2$, $26$ hours for $r=4$, and $40$ hours for $r=8$. Time to generate a super resolved image is $1$ ms for $r=2$, $1.4$ ms for $r=4$, and $1.9$ ms for $r=8$. Training and test was performed on a NVIDIA Tesla K$40$ GPU with $12$ GB RAM. 

\begin{table*}[t]
\begin{tabular}{|c|c|c|c|c|c|c|c|c|c|c|c|}
\hline
{} & \multicolumn {5}{|c|}{Scaling factor(r) = $4$} & \multicolumn {5}{|c|}{Scaling factor (r) = $8$} & {r=16} \\  \hline
{} & {SSIM} & {RMSE}  & {PSNR} & {S3} & {$p$} & {SSIM} & {RMSE}  & {PSNR} & {S3} & {$p$} & {SSIM} \\ 
{} & {} & {($10^{-6}$)}  & {dB} & {} & {} & {} & {($10^{-6}$}  & {dB} & {} & {} & {}\\ \hline
{$SRGAN_{Sal}$} & 0.89 & {6.2} & {44.3}  & 0.83 & - & 0.84 & 7.5 & 39 & 0.74 & - & {0.80} \\ \hline
{$SRGAN_{Ledig}$} & {0.78} & {8.1} & 36.4 & 0.65 & $<0.001$ & 0.73 & 9.3 & 31 & 0.60 & $<0.001$  & {0.69} \\  \hline
{$SRCNN$}  & {0.75} & {9.1}  & 34.3 & 0.61 & $<0.009$ & 0.67 & 10.9 & 28 & 0.57 &$<0.001$ & {0.64}  \\  \hline
{SR-RF}   & {0.71} & {10.3}  & 30.2 & 0.57 & $<0.009$ & 0.62 & 12.3 & 25 & 0.55 & $<0.001$ & {0.59} \\  \hline
{SSR}  & {0.67} & {11.2}  & 27.1 & 0.54   & $<0.001$ &  0.60 & 13.7 & 22 & 0.21 & $<0.001$ & {0.56}  \\  \hline
\end{tabular}
\caption{Comparative results of different methods for image super resolution.}
\label{tab:ISR_res1}
\end{table*}

\subsection{Image Super Resolution Results}
The following ISR methods are used for comparison: 1) $SRGAN_{Ledig}:$ - the baseline GAN using MSE and CNN loss \cite{Srgan}; 2) $SRGAN_{Sal}:$ - our proposed method;
 3) $SRCNN:$- CNN based method of \cite{SRCNN}; 
 4) $SR-RF:$- random forest based method of \cite{SR-RF};
 5) $SSR:$- self super resolution method of \cite{Jog}.
Resized $1024\times1024$ images are ground truth HR images, $I^{HR}$, which  are downsampled by different $r$ to obtain $I^{LR}$ from which $I^{SR}$ are generated. 
$Y-$channel images of $I^{HR}$ and $I^{SR}$ are used to compute: 1) peak signal to noise ratio (PSNR);  2) structural similarity (SSIM) \cite{SSIM}; 3) $S3$ - the sharpness metric of \cite{S3Metric}; and 4) root mean square error (RMSE). Higher values of $1,2,3$ and lower values of $4$ indicate better performance.

\begin{figure}[t]
\begin{tabular}{cccc}
\includegraphics[height=2.9cm,width=2.9cm]{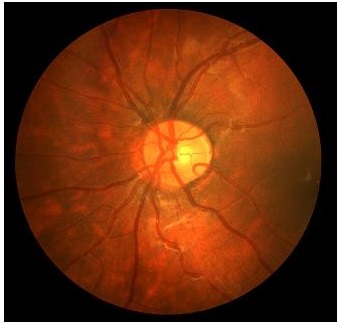} &
\includegraphics[height=2.9cm,width=3.1cm]{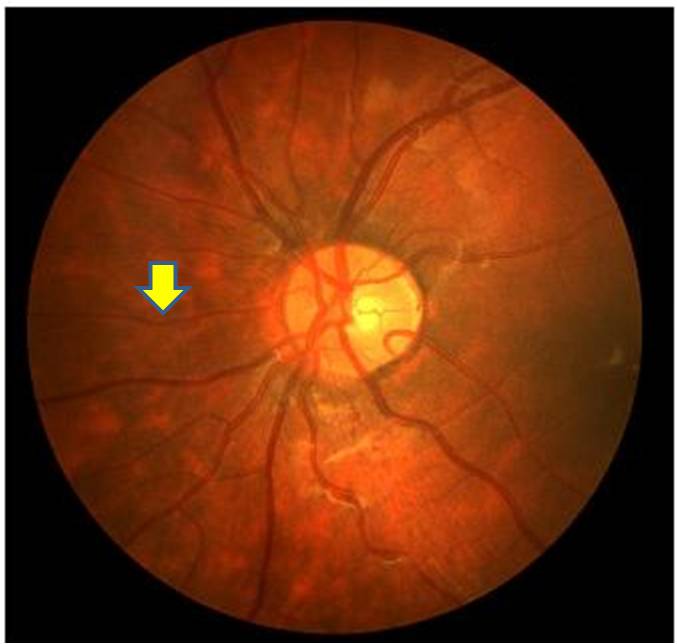} &
\includegraphics[height=2.9cm,width=3.1cm]{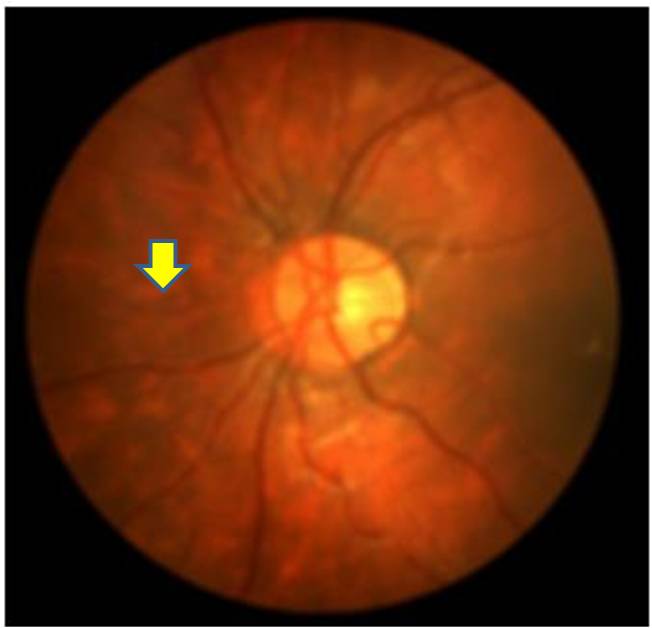} &
\includegraphics[height=2.9cm,width=2.9cm]{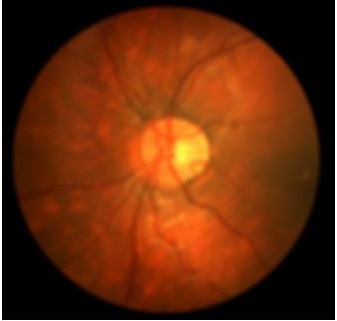} \\
(a) & (b) & (c) & (d)\\ 
\end{tabular}
\caption{Super resolution results for $r=4$. (a) original HR; SR images from: (b) $SRGAN_{Sal}$; (c) $SRGAN_{Ledig}$; and (d) $SRCNN$. }
\label{fig:SuperRes}
\end{figure}

Results of $5-$fold cross validation for $r=4,8$ are presented in Table~\ref{tab:ISR_res1}. Due to space constraints only SSIM values are shown for $r=16$.
For $r=2$, performance difference of all methods is small but becomes more pronounced for higher $r$. $SRGAN_{Sal}$ gives the best results for all $r$, and the improvement over competing methods is significant as is evident from the $p-$values of Wilcoxon signed-rank tests. 
Figure~\ref{fig:SuperRes} shows results of the top $3$ methods (due to space constraints) for $r=4$. $SRGAN_{Sal}$ shows the best performance as is evident from the SR image in  Fig.~\ref{fig:SuperRes} (b) where one of the minor retinal branches (indicated by yellow arrow) is clearly visible. On the other hand the SR image by $SRGAN_{Ledig}$ (Fig.~\ref{fig:SuperRes} (c)) is blurry and does not clearly show this retinal branch. Other methods perform much worse, with significant blur visible for the main branches as well. Clearly, $SRGAN_{Sal}$ gives the closest reconstruction to the HR image of Fig.~\ref{fig:SuperRes} (a).

\textbf{Importance of Saliency Maps}:
Excluding $l^{SR}_{CNN}$, and using $l^{SR}_{Sal}$ and $l_{w-MSE}$ for $r=4$ gives SSIM$=0.81$, RMSE$=6.9$, PSNR$=38.6$ dB, and S3$=0.69$. They are slightly higher than $SRGAN_{Ledig}$, indicating local saliency maps alone perform better than $l^{SR}_{CNN}$ and MSE in preserving image information. Combining local saliency information with CNN loss significantly improves SR image quality. Using either curvature (SSIM$=0.82$) or entropy (SSIM$=0.83$) for $r=4$ lowers performance,thus highlighting their individual importance in the final saliency map.
Using saliency maps of \cite{Perazzi_CVPR} gave SSIM$=0.80,0.74,0.70$ for $r=4,8,16$. Our proposed local saliency maps outperforms \cite{Perazzi_CVPR} (a global saliency map) as it does not capture fine structural information in retinal fundus images.

\subsection{Retinal Blood Vessel Segmentation Results}

\begin{figure*}[t]
\begin{tabular}{cccccccc}
\includegraphics[height=1.6cm,width=1.4cm]{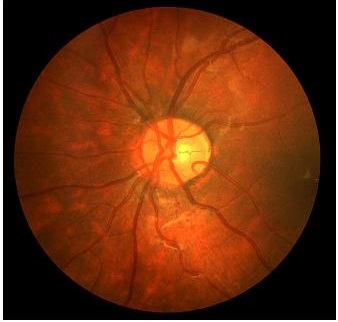} &
\includegraphics[height=1.6cm,width=1.4cm]{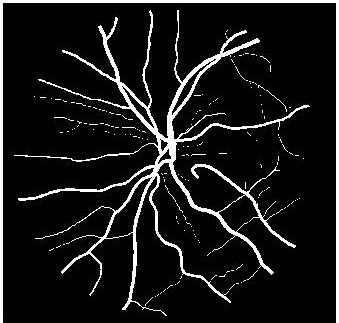} &
\includegraphics[height=1.6cm,width=1.4cm]{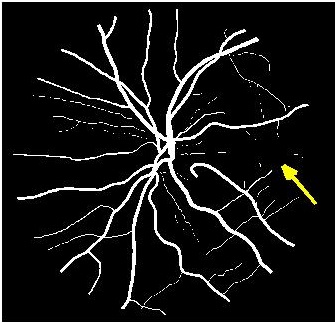} &
\includegraphics[height=1.6cm,width=1.4cm]{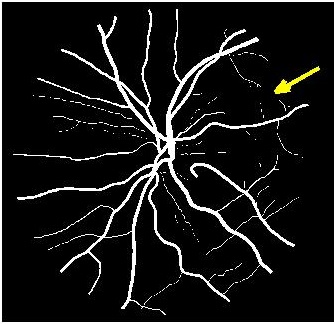} &
\includegraphics[height=1.6cm,width=1.4cm]{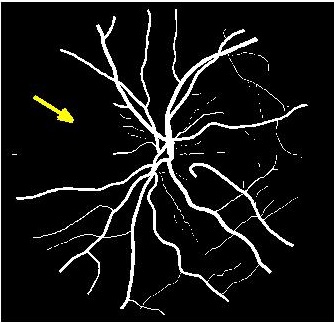} &
\includegraphics[height=1.6cm,width=1.4cm]{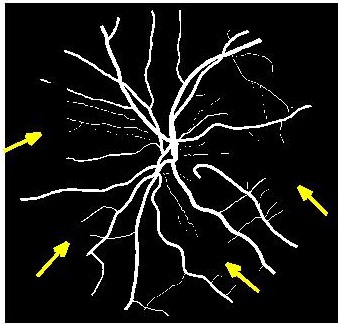} &
\includegraphics[height=1.6cm,width=1.4cm]{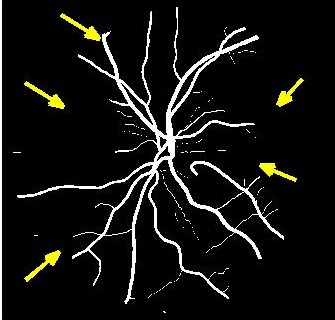} &
\includegraphics[height=1.6cm,width=1.4cm]{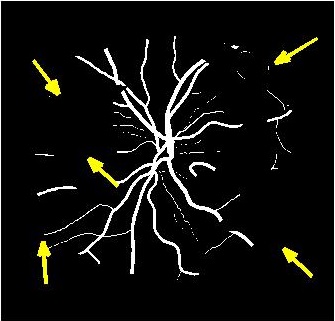} \\
(a) & (b) & (c) & (d) & (e) & (f) & (g) & (h) \\ 
\end{tabular}
\caption{Results for retinal vessel segmentation; (a) retinal image; (b) manual ground truth mask; results obtained when training on (c) orginal HR images; SR images by (d) $SRGAN_{Sal}$; (e) $SRGAN_{Ledig}$ ; (f) $SRCNN$; (g) SR-RF; (h) SSR. Yellow arrows highlight regions of inaccurate segmentation.}
\label{fig:res1}
\end{figure*}

We present retinal vessel segmentation results on the DRIVE \cite{DRIVE}, STARE \cite{STARE} and CHASE$\_$DB1 \cite{ChaseDB1} datasets with $40$, $20$ and $28$ images respectively. 
Original images and manual annotations ($I^{HR}$) are downsampled by $r=4,8$ to get $I^{LR}$ and $5$ sets of $I^{SR}$ from $5$ methods trained on \cite{data}. 
$I^{HR}$ and the $5$ sets of $I^{SR}$ were used to train $6$ different state-of-the art U-Nets for vasculature segmentation \cite{Ret_Unet}. 
The average accuracy ($Acc$) and sensitivity ($Sen$) for $r=4,8$ is summarized in Table~\ref{tab:SrRes_vasc}. Better ISR methods should give higher vessel segmentation accuracy and performance of $I^{HR}$ gives a lower bound on the segmentation error. 
$SRGAN_{Sal}$'s performance is closest to $I^{HR}$, and establishes its superiority over all competing methods. 
Figures~\ref{fig:res1} (a)-(h) show results of vessel segmentation on an example image. $SRGAN_{Sal}$'s performance is most similar to $I^{HR}$ as is evident from the areas of inaccurate segmentation highlighted by yellow arrows. Most of the methods do not segment the finer vasculature structures, while SSR and SR-RF are unable to segment some of the major arteries.

\begin{table}[t]
\begin{tabular}{|c|c|c|c|c|c|c|c|c|c|c|c|c|}
\hline
{} & \multicolumn {2}{|c|}{HR} & \multicolumn {2}{|c|}{$SRGAN_{Sal}$} & \multicolumn {2}{|c|}{$SRGAN_{Ledig}$} & \multicolumn {2}{|c|}{$SRCNN$} & \multicolumn {2}{|c|}{SR-RF} & \multicolumn {2}{|c|}{SSR} \\  \hline
{} & {Acc} & {Sen}  & {Acc} & {Sen} &{Acc} & {Sen} & {Acc} & {Sen}  & {Acc} & {Sen} &{Acc} & {Sen} \\ \hline
{DRIVE} & 0.98  & 0.79  & {0.96} & {0.77} & 0.92  & 0.74 & 0.89 & 0.73 & 0.87  & 0.70 & 0.85 & 0.69 \\  \hline
{STARE} & 0.98  & 0.90  & 0.96 & 0.87 & 0.91  & 0.84  & 0.89 &  0.81 & 0.86  & 0.77  & 0.83 & 0.72 \\  \hline
{CHASE$\_$DB} & 0.97 &  0.84  & 0.95 & 0.82  & 0.91  & 0.76  &  0.87 &  0.72& 0.85 &  0.70  & 0.81 & 0.68  \\  \hline
\end{tabular}
\caption{Comparative vasculature segmentation results of different super resolution methods. The values are for scaling factor $4$ and $8$.}
\label{tab:SrRes_vasc}
\end{table}

\section{Conclusion}
\label{sec:concl}

We have proposed a novel method for super resolution of retinal fundus images based on GANs. Local saliency maps effectively  quantify a pixel's perceptual relevance, and are used to weight each pixel according to its importance and define a novel saliency loss. When incorporated into the GAN loss function, the resulting SR images are better than those obtained using CNN feature loss. Experimental results show combination of saliency and CNN loss significantly outperforms current state of the art GANs and other competing ISR methods. 
The resulting super resolved images can be used to increase the size and resolution of low dimensional images, and then apply different image analysis algorithms. When using our SR images for retinal vessel segmentation the results are close to those obtained with the original high resolution images. Our method can be applied for other medical images as well.

\bibliographystyle{splncs03}
\bibliography{MICCAI_SuperRes_Ref}

\end{document}